\documentclass[11pt, a4paper]{article}

\usepackage[margin=1in]{geometry}
\usepackage{graphicx}
\usepackage{authblk} 
\usepackage{color}
\usepackage{dirtree}
\usepackage{url}
\usepackage{eurosym}
\usepackage{rotating} 

\usepackage{amssymb}
\usepackage{pifont}
\usepackage{amsmath}
\usepackage{multirow} 
\usepackage{hyperref}

\newcommand{\cmark}{\ding{51}}%
\newcommand{\xmark}{\ding{55}}%

\title{Still image and spatial-temporal tomato data enabling detection, segmentation, tracking, and video-instance segmentation using strong and weak labels}
\author[1,3]{Michael Halstead}
\author[1,3]{Esra Guclu}
\author[1]{Mohamed Farag}
\author[1]{Enrico Pallotta}
\author[1]{Christian Hund}
\author[1]{Ribana Roscher}
\author[1]{Maren Bennewitz}
\author[1]{Juergen Gall}
\author[1]{Cyrill Stachniss}
\author[1,2]{Chris McCool} 

\affil[1]{University of Bonn, Bonn, Germany}
\affil[2]{CSIRO, Pullenvale, Australia}
\affil[3]{These authors contributed equally to this manuscript}
\date{\today} 

\begin{document}

\maketitle

\begin{abstract}
In this manuscript we release two datasets for visual sensing of tomato plants grown in commercial-like settings and acquired using a robot.
The first is BUTom21 which consists of still images and manual annotations.
The second is BUTom-ST21 which consists of video-based data and semi-automated annotations through AI-based methods, referred to as pseudo-labels.
In both cases, we provide pixel-level labels for the ripeness of the fruit.
The aim is to provide the research community a challenging set of real-world imagery to explore methods to sense and estimate the state of tomato plants and their fruit, which is an important horticultural crop.
Importantly, the spatial-temporal dataset provides individual fruit count and ripeness information enabling researchers to push the boundaries of field-based phenotyping.
\end{abstract}

\section{Background \& Summary}

Horticultural robotics is fast becoming an important part of society through sustainable farming.
With the ever-increasing need for food production as the population grows, the reduction in arable farmland, and an aging workforce~\cite{saizrubio2020, yepezponce2023}, robotics is quickly becoming an accurate and efficient solution to address these issues~\cite{barbosa2024, yepezponce2023}.
A stymying factor in the uptake of robotics and robotic vision in industry is the paucity of available data to train machine learning based algorithms for various tasks, such as monitoring~\cite{smittPathobot2021}, planning~\cite{Josegovmp2025}, or harvesting~\cite{arad2020development}.
To address the issue of data paucity, we recently proposed an approach to obtain pseudo-labels for video data by using a neural radiance-field approach~\cite{smitt2022pag} and applied this to sweet pepper grown in glasshouses to produce the BUP-ST20~\cite{2025bupst20} dataset. 
In this manuscript, we expand beyond sweet pepper and include tomato which is a key agricultural crop.
We do this by releasing two novel tomato datasets: BUTom21 and BUTom-ST21. 

Tomatoes are an important crop in today's broader social context.
In Europe alone, the tomato harvest in 2024 was estimated to be 16.9 million tonnes of produce~\cite{eurostat2025crops}, which was a considerable increase of 5.7\% from 2023.
Currently, tomatoes make up almost 20\% of all vegetable crops grown globally~\cite{cbi2023tomato}, making it one of the most commonly grown vegetables in the world.
Due to a cultural shift in farming and the increase in farming costs Europe imports approximately \euro{}5 billion of tomatoes every year~\cite{cbi2023tomato}.
For farming these fruit, the labor expense can be prohibitive and robotic platforms are currently being developed to reduce costs and improve efficiency of farming procedures.

Robotics-based technological advances are evident in a number of fields, not just agriculture and horticulture.
Data-driven agricultural robotics approaches rely heavily on the efficacy of their underlying models.
In general, models and techniques are developed for a specific problem in a specific setting.
In~\cite{halstead2021agnostic} it was shown that models can be adapted to both horticulture and agriculture with only a change in input data (arable farmland versus glasshouses).
However, to be able to achieve accurate and sustainable robotic vision approaches extensive data is required.
When we consider foundational models such as SAM3~\cite{CarionSam3} we see that, given enough generalized data, instance-based semantic segmentation is possible.
The problem lies in the available data, particularly in agriculture and horticulture.

To date, most vision-based agricultural datasets can be considered small when compared to those such as ImageNet~\cite{deng2009imagenet} or COCO~\cite{lin2014microsoft}.
This creates significant gap between supply and demand in the robotic vision and artificial intelligence fields in agriculture.
In order to bridge this widening gap more accurate and robust data is required, particularly for instance-based semantic segmentation tasks.
Instance-based semantic segmentation is an important component within agriculture as it provides high level information that enables a number of other important tasks.
It not only infers where each object (instance) is in an image but also all the pixels associated with each object.
This enables downstream tasks which would not be possible without object-level pixel-wise masks and the fruits or vegetables respective ripeness estimation.

Currently, in agriculture and horticulture, the datasets for instance-based semantic segmentation are small, containing as few as 82 images in the training set~\cite{afonso2020tomato}.
Recently, datasets such as BUP-ST20~\cite{2025bupst20} and APPLE MOTS~\cite{de2022apple} incorporate more frames in their training set, allowing more generalized learning; these two datasets were developed for spatial-temporal tasks and not specific to still image segmentation.
However, most common horticultural datasets are aimed at crops such as sweet peppers~\cite{smittPathobot2021, 2025bupst20}, oranges~\cite{james2024}, berries and peaches~\cite{wang2025}, and apples~\cite{hani2020}.
This is in direct contrast to our earlier statistic that tomatoes are one of the most commonly produced vegetable crop in the world.

Recently, there has been a push to include more tomato-based datasets.
In 2020~\cite{tsironis2020tomatod} released tomatoOD, a detection-based tomato dataset with 277 images in the training set and 55 images in the evaluation set.
Unfortunately, from a robotics perspective, this dataset removed tomatoes that were occluded more than 50\% of their appearance or if the fruit was too small or blurred.
Also in 2020, Rob2Pheno~\cite{afonso2020tomato} was released and captured in a glasshouse on RealSense D435 cameras. 
This is a small dataset, containing only 123 images in total (82 train/ 41 evaluation) and only the tomatoes in the foreground were annotated. 
The main benefit of this dataset is that they released instance-based semantic segmentation labels for all foreground tomatoes in the scene. 
With 804 images (643 train/ 161 evaluation) in its corpus Laboro Tomato~\cite{laborotomato2023} is one of the biggest tomato datasets aimed at segmentation and detection.
It is captured on two different resolution cameras ($3024\times{4032}$, $3120\times{4160}$) which adds further complexity to an already complex task.
Their dataset also has the benefit of capturing multiple cultivar of tomatoes, both normal sizes and cherry tomatoes; and labels the subclasses into six different classes.
In general, this dataset is not considerably complex in terms of the images, the tomatoes are assured to be the focus of the image and the full plant/vine/truss is not always captured in the image.
Finally, in 2026 Zhang et~al.~\cite{zhang2026tomato} released both TomatoMAP-Det and TomatoMAP-Seg, as subsets of TomatoMAP.
TomatoMAP-Det is a detection-based dataset with approximately 64k images.
The segmentation dataset (TomatoMAP-Seg) contains 3,616 images, making it the largest known tomato dataset for segmentation purposes.
This dataset also comes with 10 subclasses ranging from flower stage through to the fully ripe stage.
In their approach, they leveraged SAM2~\cite{ravi2025sam} to get the pseudo-labels of their dataset and then hand-corrected the ground truth, creating a known quantity to evaluate on.
In general, tomato datasets are relatively sparse in the research community; we help to address this gap by introducing our two novel tomato datasets.

For sweet pepper, a well-represented fruit in research, and another highly produced fruit in agriculture, there are a number of well-known instance-based semantic segmentation datasets. 
Early datasets such as~\cite{halsteadbup19, smittPathobot2021} are small in stature but contain pixelwise labels for all fruit in a glasshouse.
In~\cite{halsteadbup19} the images are captured on a hand-held camera scanning rows in a glasshouse, while~\cite{smittPathobot2021} uses the robotic platform PATHoBot to capture all rows in a glasshouse cell.
The illumination in these datasets, and many robotic vision datasets for agriculture is another concern.
They are heavily impacted by the time of the day of capture (the position of the sun), the inclusion of the sun shield over the glasshouse, and the direction of the robot.
To alleviate some of these concerns the WUR dataset~\cite{barth2018} exploits artificial light to maintain more consistency within the dataset.
However, unless the robotic platform of capture consistently run at night with artificial lighting this is less generalized to the horticultural domain.
Moving away from sweet pepper, one of the largest instance-based semantic segmentation datasets for fruit is MegaFruits~\cite{wang2025}.
MegaFruits contains over 25k images of strawberries, blueberries, and peaches fully annotated with pixel-wise masks.
In this dataset, the concentrate on the important task of occlusion and densely clustered fruit.
Another large fruit-based dataset is MinneApple~\cite{hani2020}, which contains approximately 1k images with over 41k pixel-wise masks.
In their dataset they again concentrate on occlusion of apples as they are generally densely clustered.
Overall, these datasets show a recent push to include large amounts of fruit and vegetable data for the research community.
However, these concentrate on still image information, meaning spatial-temporal models can not be trained as consistent tracklet IDs are not included.

For research on video instance segmentation (VIS) and multi-object tracking (MOT) very limited data is available in the agriculture and horticulture domains.
We recently released BUP-ST20~\cite{2025bupst20} which is applied to sweet pepper and leverages the power of pseudo-labels generated through a neural radiance-field approach.
As far as we are aware, this is still the largest dataset available that was captured on a robotic platform, contains instance-based semantic segmentation, robot odometry, depth images, and color (RGB) images in its corpus.
Likewise, APPLE MOTS~\cite{de2022apple} was released to the public in 2022.
Similarly to sweet peppers and tomatoes, it is based on the principle that common tracking datasets contain objects (such as pedestrians dressed in different clothing) which are somewhat easier to disambiguate.
For horticulture, fruit is homogeneous in nature, generally speaking red tomatoes are harder to tell apart and re-identify at a later date.
While, the different subclasses can be told apart from each other (orange versus green versus red) the same subclass can be much more difficult to distinguish.

To help alleviate these gaps, homogeneous VIS and MOT datasets and dataset paucity, we introduce the BUTom21 and BUTom-ST21 datasets.
We follow the same protocols as our previous datasets, BUP20 and BUP-ST20, using similar techniques to deliver the final datasets.
To download the data and see the minimal working code for running the dataloaders please see the github repostiory \url{https://github.com/Agricultural-Robotics-Bonn/BUTom21-ST21}.
The datasets capture the tomato cultivar \textit{sweeterno} from early stages of ripeness after full flowering, through to harvestable fruit. 
The two datasets are developed for different purposes and with different challenges and limitations; however, the main purpose revolves around instance-based semantic segmentation.

In more detail, the BUTom21 dataset is a fully hand-annotated still image dataset of tomatoes.
It contains 123, 72, 98 images respectively in the training, validation, and evaluation subsets, and all annotations are supplied in a Coco format based json file.
The dataset contains RGB and noisy depth images captured across 2 capture days and two different RealSense D435i cameras.
In contrast to other datasets, and similar to BUP20, we annotate all tomatoes in the scene, not just those in the foreground.
This creates a fully pixel-wise annotated dataset for tomato detection and segmentation in a real-world robotics scenario.

The BUTom-ST21 alleviates the data paucity issue within the horticulture and tomato domain.
In total we have 7749, and 4536 pseudo-labeled images across the training and validation set, and 1386 images hand-labeled in the evaluation set.
To create the dataset we follow a similar design paradigm to the BUP-ST20 dataset.
The inclusion of tracklet IDs enables VIS and MOT evaluation in a highly dynamic horticultural scene with homogeneous fruits.
A key difference from the BUP-ST20 dataset was the lack of robot odometry information. 
In BUTom-ST21 we did not have the benefit of a working odometry and therefore had to revert to traditional approaches to provide camera pose information. 
This paradigm shows that spatial-temporal datasets can be created when there is smooth movement of the camera(s).

Our two released novel datasets, BUTom21 and BUTom-ST21 are freely available for resarch purposes.
We have separated these two datasets into two explicit locations to facilitate easier downloading and task specific usage.
The BUTom21 dataset is a fully hand-labeled tomato dataset with subclass assignment.
The training and validation subsets from BUTom-ST21 are pseudo-labeled using the BUTom21 dataset as the training information, and a fully hand-corrected evaluation subset.
Overall, we created two complex novel datasets for various real-world agricultural robotic vision purposes.
In this manuscript we:
\begin{enumerate}
    \item Release a fully hand-annoated still image tomato dataset;
    \item Release a spatial-temporal tomato dataset with pseudo-labeled training and validation subsets and hand-corrected evaluation subset; and
    \item Provide extensive benchmarking of the two datasets for various tasks (outlining the challenges and limitations of each).
\end{enumerate}

\section{Methods}

For our two novel datasets, BUTom21 and BUTom-ST21, the following section describes the data collection procedure and the various annotation tasks undertaken to create both the still image dataset (BUTom21) and the spatial-temporal version (BUTom-ST21).
The data capture happened at the University of Bonn's commercial-like glasshouse setup at campus Klein-Altendorf.
Multiple crops are grown in the multi-cell glasshouse and of interest for this dataset was the \textit{sweeterno} tomatoes grown in one such cell alone (no other fruit or vegetables are mixed with the tomatoes).
The glasshouse cell itself is arranged such that there are six rows, where the two outside rows have sterile backgrounds and each row is about 36m in length.
To capture the data we employed the PATHoBot~\cite{smittPathobot2021} robotic platform which exploited the piperails installed parallel to the cropping rows.
The piperails and the use of PATHoBot ensured we have consistent distance from the cameras to the heating rails in the rows (approximately 1.2m).
PATHoBot is designed with four RealSense D435i cameras mounted evenly spaced along a vertical pole, this ensures that the entire tomato vine can be captured in one scan.
For these datasets we utilized the two middle cameras, the topmost camera was deployed too high and thus did not capture any tomatoes, and the lowest camera captured superfluous information such as potting equipment and the ground.
The RealSense D435i capture both RGB and registered depth information at a resolution of $1280\times720$ and at approximately 15Hz; further details on PATHoBot and the camera setup can be found in~\cite{smittPathobot2021}.

The \textit{sweeterno} is a medium to large sized vegetable that ripens in five distinct stages: green, mixed red, red, mixed orange, and orange.
To ensure tomato subclass (ripeness) diversity, we captured all six rows on two separate occasions, approximately one month apart.
By utilizing the two middle cameras we further added variety to the subclasses as more ripe tomatoes (red) appear at the the bottom of the vine, and more juvenile (green) appear towards the top.
This is in contrast to the sweet pepper datasets BUP20 and BUP-ST20 which only uses a single camera.

While our two publicly released datasets, BUPTom21 and BUTom-ST21, are derived from same images they are processed differently and released separately.
BUTom21 is a fully hand-labeled dataset for still image tasks such as detection and segmentation.
BUTom-ST21 is a much larger spatial-temporal dataset with pseudo-labeled training and validation subsets and fully hand-corrected evaluation subset.

The still image-based BUTom21 dataset contains RGB and depth images from the two cameras captured over the two days.
We uniformly select and then manually refine the selected images to be annotated.
Our manual refinement was undertaken to ensure dataset diversity as well as equal camera and capture day representation.
Image selection was also done to ensure the dataset was diverse enough to fully encapsulate the various challenges witnessed during capture.
One of the key challenges witnessed during capture was the inclusion of ``tiny'' objects. 
These ``tiny'' objects occur due to juvenile tomatoes and largely occluded tomatoes; occlusion can occur because of other tomatoes, leaves, stems, trusses, and other objects.
Other challenges witnessed during data capture were camera blurring due to vibration, illumination variation, scene complexity (inner versus outer rows), and subclass distribution; each of these variations was represented explicitly in the images annotated in the dataset.
To ensure fairness of results and hyper-parameter tuning, we define three subsets of fully annotated data. 
There is a total of 293 images in the dataset, with 123, 72, and 98 images respectively for the training, validation, and evaluation subsets.
The images for each subset were also extracted from specific rows in the glasshouse cell.
The training subset is derived from rows 1 and 2 on the first day and rows 1, 2, and 3 on the second day of capture.
While validation used row 3 and 4 on the first day and then only row 4 on the last day, and evaluation is composed of rows 5 and 6 on both days; ensuring a complete hold out and non-overlapping evaluation set.

\begin{figure}[ht]
    \includegraphics[width=0.95\linewidth]{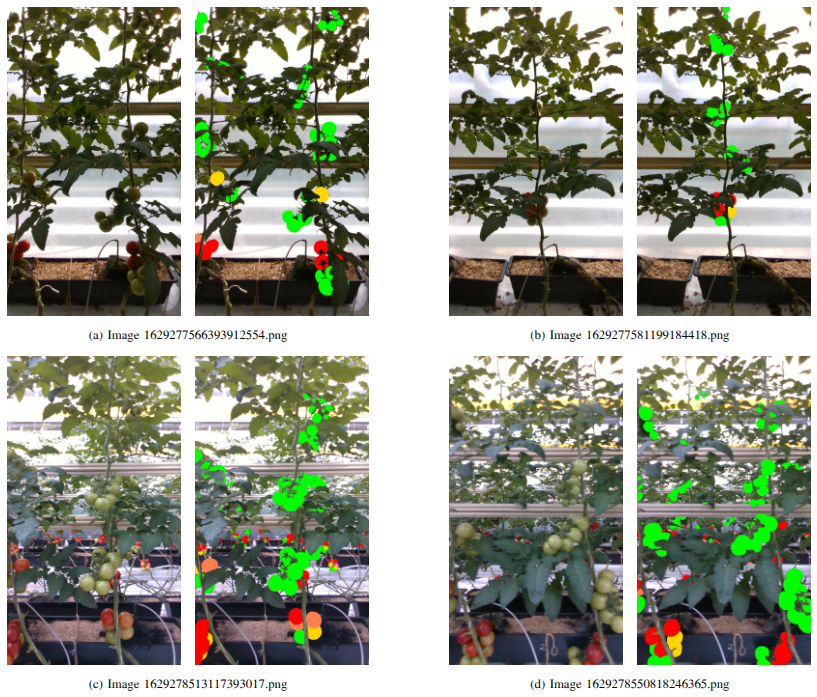}
    \caption{Example images of the still image-based BUTom21 dataset. (a)-(b) These two images have a sterile background with no tomatoes in deeper rows, also, (b) has very limited tomatoes in the scene. (c)-(d) Complex background with a number of tomatoes seen in deeper rows. The difference between (a)-(b) and (c)-(d) is also the illumination differences.}
    \label{fig:butom21_ex}
\end{figure}

The annotation process used a single person per image to annotate all tomatoes in the scene (foreground and background) with consistency checks by another operator.
Tomatoes are annotated with individual pixel-wise masks and their associated subclass (green, mixed red, red, mixed orange, and orange) and stored in a json file in the Coco format using the Coco annotation software~\cite{cocoannotator}.
In \url{https://github.com/Agricultural-Robotics-Bonn/BUTom21-ST21} we include a minimum working example of how to extract the images in a PyTorch style dataloader.
Table~\ref{tab:bdbuptom21} describes the full breakdown of the BUTom21 dataset, including the occurrence of each subclass and their associated labels in the json file.
Overall, BUTom21 provides a new tomato dataset which can be used for any task that requires instance-based semantic segmentation labels, and examples of the dataset are shown in Figure~\ref{fig:butom21_ex}.

\begin{table*}[hb]
    \caption{Breakdown of the distribution in the BUTom21 and BUTom-ST21 datasets for still images and spatial-temporal video sequences. BUTom21 is at the top 3 rows, and BUTom-ST21 is the following 10 rows. For the BUTom21 dataset we also include the associated subclass (ripeness) labels that are in the json file.}
    \label{tab:bdbuptom21}
    \resizebox{\textwidth}{!}{
    \begin{tabular}{|l|c|c|c|c|c|c|c|c|c|c|}
        \hline
        Still Image & Images & Red (0) & Mixed Red (1) & Green (2) & Orange (3) & Mixed Orange (4) & Tiny & Small & Medium & Large \\
        \hline
        Training & 123 & 1268 & 40 & 6189 & 211 & 317 & 1479 & 6355 & 191 & 0 \\ 
        Validation & 72 & 963 & 107 & 4503 & 186 & 292 & 1130 & 4772 & 149 & 0 \\ 
        Evaluation & 98 & 1011 & 20 & 4367 & 265 & 278 & 1205 & 4570 & 166 & 0 \\ 
        \hline
        \hline
        Video Sequences & Images & Red & Mixed Red & Green & Orange & Mixed Orange & Tiny & Small & Medium & Large \\
        \hline
        Training(m2f) & 7749 & 24383 & 622 & 161065 & 4804 & 7970 & 1139 & 186420 & 11285 & 0 \\
        Training(yolo) & 7749 & 29173 & 509 & 175675 & 2757 & 8446 & 1331 & 201602 & 13627 & 0 \\
        Validation(m2f) & 4536 & 14067 & 571 & 85869 & 2483 & 7308 & 834 & 100669 & 8795 & 0 \\ 
        Validation(yolo) & 4536 & 17173 & 476 & 92323 & 755 & 6446 & 708 & 105902 & 10563 & 0 \\ 
        Evaluation & 1386 & 4723 & 300 & 35807 & 1114 & 1946 & 804 & 40489 & 2597 & 0 \\
        \hline
        Video Sequences & Tracklets & Red & Mixed Red & Green & Orange & Mixed Orange & Tiny & Small & Medium & Large \\
        \hline
        Training(m2f) & 7768 & 974 & 23 & 6190 & 186 & 295 & 64 & 6962 & 642 & 0 \\
        Training(yolo) & 8311 & 1153 & 20 & 6727 & 93 & 318 & 48 & 7508 & 744 & 0 \\
        Validation(m2f) & 4625 & 593 & 22 & 3605 & 105 & 300 & 50 & 4040 & 535 & 0 \\ 
        Validation(yolo) & 4869 & 718 & 21 & 3850 & 36 & 244 & 30 & 4219 & 620 & 0 \\ 
        Evaluation & 1655 & 186 & 8 & 1343 & 45 & 73 & 6 & 1478 & 171 & 0 \\
        \hline

    \end{tabular}}
\end{table*}

The spatial-temporal BUTom-ST21 dataset is derived from the same images as BUTom21, and relies heavily on the original still image dataset.
It contains 123, 72, and 22 sequences, respectively, for the training, validation, and evaluation subsets; Table~\ref{tab:bdbuptom21} breaks down the dataset.
Similar to BUP-ST20 we exploit pseudo-labels in the training and validation subsets, and hand-corrected annotations in the evaluation set.
In total we have 7749 and 4536 images with pseudo-labels (training and validation) and 1386 fully annotated images.
As a point of difference to BUP-ST20, which has a random number of frames per sequence, we ensure that each sequence has a consistent number of frames which is 63.
As this is a spatial-temporal dataset, it is designed to be used for VIS and MOT purposes and so each pixel-wise annotation also includes a tracklet identifier.
This ensures (or weakly ensures for the training and validation subsets) that each object in the scene is tracked through the sequence and obtains a unique ID per sequence.

To obtain our pseudo-labels we use PAg-NeRF~\cite{smitt2022pag} which is a neural radiance field technique (NeRF).
PAg-NeRF takes as input RGB images, corresponding depth images, instance masks generated from depth filtered versions of BUTom21, and camera pose information for each frame.
To generate each sequence, first we position an image from BUTom21 in the middle of a sequence and then we extract a total of 83 frames around it (approximately 41 each side), ensuring no overlap between sequences.
We extract 83 frames as PAg-NeRF needs ``warm up'' frames to visualize the scene correctly at our first actual starting frame.
The actual images annotated in sequences (63 images) start from the tenth image of the extracted images, and ends at the 73rd image.
The instance-based semantic segmentation has been extracted by two well known techniques: Mask2Former~\cite{chengm2f2022} and Yolo26~\cite{jocher2026ultralyticsyolo26unifiedrealtime} to compare the performance of both approaches.
In our released data we have included the pseudo-labels from both approaches.

As we are only concerned with tomatoes in the current row we are scanning, the first step before training our two techniques is depth filtering.
Due to the manner in which tomatoes grow and the noise in the depth images, we select a depth of 1.4m from the camera (compared to 1.2m in~\cite{2025bupst20}).
We then take the annotated pixel-wise mask from BUTom21, and if at least 50\% of the mask is contained within $0\rightarrow{1400}$ from our depth map we keep the mask, otherwise we remove it from the training procedure.

Once we have filtered out the masks based on their depth information we can now train our instance-segmentation models, both Mask2Former and Yolo26 are trained on the depth filtered BUTom21 dataset.
Mask2Former is fine-tuned with the Swin-L~\cite{liu2021swin} backbone and initialized with Coco pretrained weights~\cite{lin2014microsoft}.
For training, we maintain the default configuration and only modify the base learning rate to be 0.00005.
We train the model for 510 epochs (with the best model selected based on the validation set), with a total batch size of 32 on 8 NVIDIA A100 GPUs. 
For Yolo26 we fine-tune using all their in-built configuration parameters, and the Ultralytics (\url{https://github.com/ultralytics/ultralytics}) internal model \textit{yolo26x-seg.pt}.
The batch size is set to 16 and evenly split across 8 NVIDIA RTX A6000 GPUs.
Images are processed at their original $1280\times{720}$ size, and both techniques use the full subclass (multi-class) information during training.
We are now able to run inference on the BUTom-ST21 sequences and produce all associated instance-based semantic segmentation masks.
It should be noted that at this point all masks are processed independently, and there is no consistent tracklet ID information.

The final step before employing PAg-NeRF for our consistency rendering, is to complete the camera pose information.
After capture it was noted that the odometry information from PATHoBot was not functioning correctly, and for some rows there was no odometry information.
To address this issue and show how new datasets can be achieved without noisy odometry information (such as that used in~\cite{2025bupst20}) we exploit structure-from-motion (SfM) techniques to compute the poses.
We also modified PAg-NeRF slightly to ensure it could handle this information directly.
For each sequence, we first run COLMAP~\cite{schonberger2016structure}, configured with the PINHOLE camera model and initialized with our known camera intrinsics to enhance the accuracy of the recovered poses.
This approach was successful in all but 13 sequences.
In the unsuccessful sequences we employed hierarchical localisation~\cite{sarlin2019coarse} (HLoc).
This approach used SuperPoint~\cite{detone2018superpoint} for feature extraction and SuperPoint+LightGlue~\cite{lindenberger2023lightglue} for matching; and was able to successfully register the remaining sequences.
For dataset completeness we also release both the HLoc and COLMAP poses for their respective sequences.

\begin{figure}
    \centering
    \includegraphics[width=0.7\linewidth]{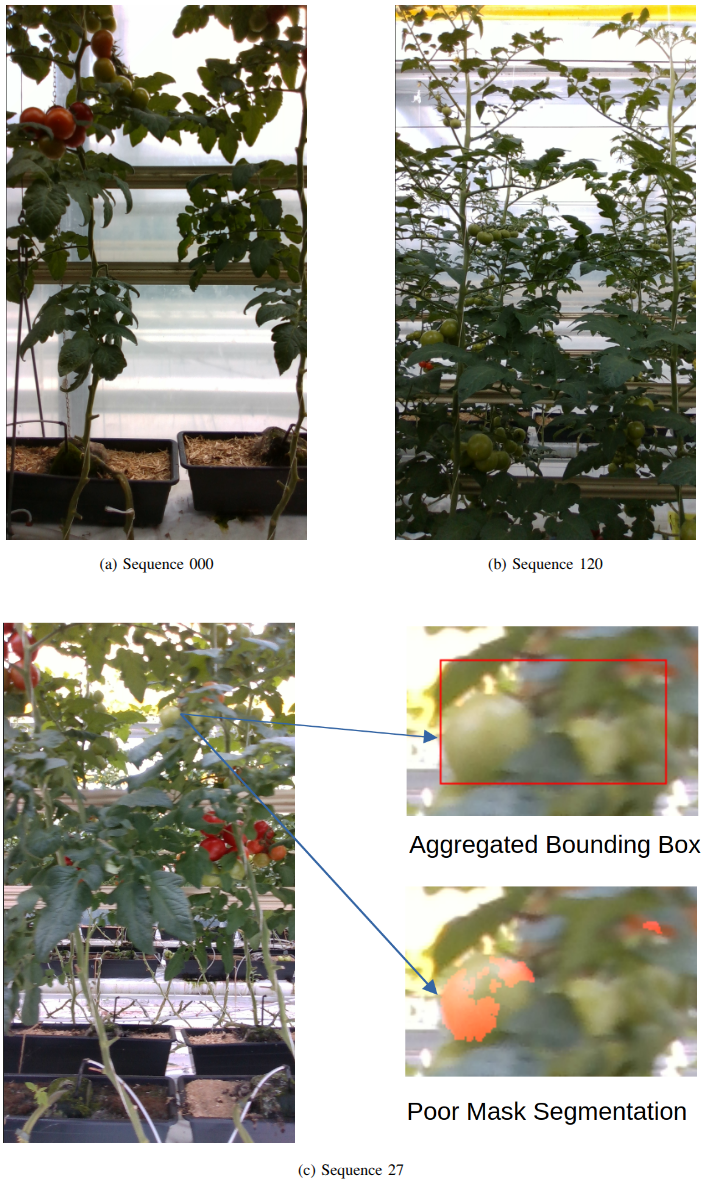}
    \caption{The issues and challenges associated with the BUTom-ST21 dataset. (a) Indicates a sterile background with low illumination. (b) Shows a complex background where only the foreground objects are annotated, and higher illumination; it should be noted that illumination is even higher in some sequences. (c) Examples of where the bounding boxes fail, where two objects are aggregated together; also, for the same object, we see the poor segmentation performance indicative of both Yolo26 and Mask2Former.}
    \label{fig:butomst21_ex}
\end{figure}

To render the consistent tracklet IDs we employ PAg-NeRF, with all of our now completed sequence information.
PAg-NeRF is trained using the default configuration parameters, except for changing the scene scale to 0.02 and the deactivation of the pruning operation during training.
We found that in both cases, by changing these values, we achieved better rendering results.
This approach achieves our first pass at pseudo-labels for all three subsets where we now have: RGB images, registered depth, instance-based semantic segmentation masks, consistent tracklet IDs, and camera poses.
For the training and validation set we do further post-processing to achieve more robust pseudo-labels due to four main issues with the original PAg-NeRF output: horizontally joining two (or more) objects, vertically joining two objects, tracklet ID switching, and finally reproducing the same tracklet ID after the first one has left the scene.
We found this post-processing approach improved AP results by a minimum factor of 4.
However, this still leaves the problem of missing small objects, and poor segmentation masks, making BUTom-ST21 a challenging dataset on which to achieve strong results.
Overall, while we have removed a number of issues, as seen in Figure~\ref{fig:butomst21_ex} there are still considerable limitations with the pseudo-labels.

Finally, for the evaluation set, we take the pseudo-labels and hand-correct them.
This is completed by human operators removing joined bounding boxes, creating consistent tracklet IDs, adding missing tomatoes, removing false positives, and fixing segmentation masks.
A final check is done by a single operator to maintain consistency.
This creates the largest pseudo-labeled and hand-labeled spatial-temporal dataset for VIS and MOT purposes using tomatoes.

\section{Data Records}

Our two novel datasets released for research purposes can be found at the github repository \url{https://github.com/Agricultural-Robotics-Bonn/BUTom21-ST21} with download instructions.
Each dataset is self contained and can be downloaded without the other.
It should be noted that the BUTom21 dataset contains only the annotated RGB and depth images in png format from two cameras.
The BUTom-ST21 dataset contains RGB and depth images for the respective sequences in the dataset along with their associated annotations.

\begin{figure}
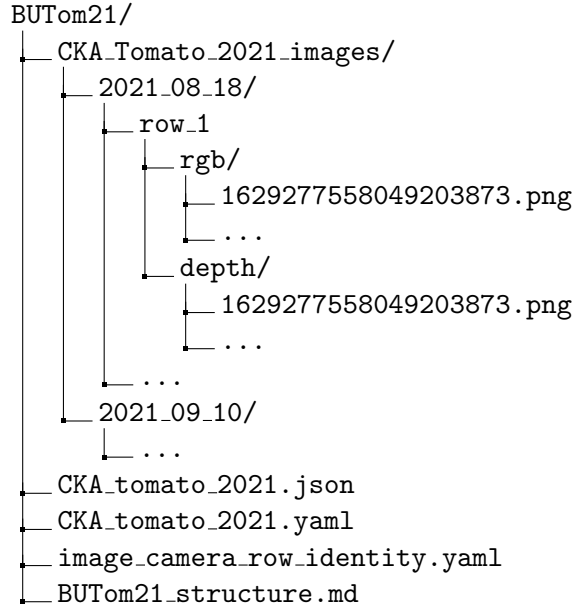

\begin{center}
\begin{minipage}{0.45\textwidth}
\dirtree{%
.1 BUTom21/.
.2 CKA\_Tomato\_2021\_images/.
.3 2021\_08\_18/.
.4 row\_1.
.5 rgb/.
.6 1629277558049203873.png.
.6 ....
.5 depth/.
.6 1629277558049203873.png.
.6 ....
.4 ....
.3 2021\_09\_10/.
.4 ....
.2 CKA\_tomato\_2021.json.
.2 CKA\_tomato\_2021.yaml.
.2 image\_camera\_row\_identity.yaml.
.2 BUTom21\_structure.md.
}
\end{minipage}
\end{center}
\caption{The file structure of the BUTom21 dataset.}
\label{fig:butom21struct}
\end{figure}

As a still image dataset, BUTom21 is separated in the manner described in Figure~\ref{fig:butom21struct}.
The main root directory contains the \textit{CKA\_tomato\_2021.yaml} file which breaks down a number of statistics as well as the subclass labels.
Most importantly, it contains the breakdown of subset identifiers in the Coco based json file.
The \textit{CKA\_tomato\_2021.json} file contains the image information and the Coco format annotations, including bounding box locations, Coco mask information (polygons), and subclass classification.
In the json, for the image information, the relative path is included, which means no extra search is required to find the desired image.
The \textit{image\_camera\_row\_identity.yaml} supplies extra information about the images, including which camera they came from and which row they were captured in.
Finally, the \textit{BUTom21\_structure.md} contains the full description of the yaml and json files, and in particular, how to use the information contained within.

From a file structure standpoint we have organized the dataset in a similar fashion to other still image-based vegetable datasets.
The depth and RGB images are organized such that they are located based on their capture date and the row they were obtained from.
An example of this breakdown is provided in Figure~\ref{fig:butom21struct}.

\begin{figure}
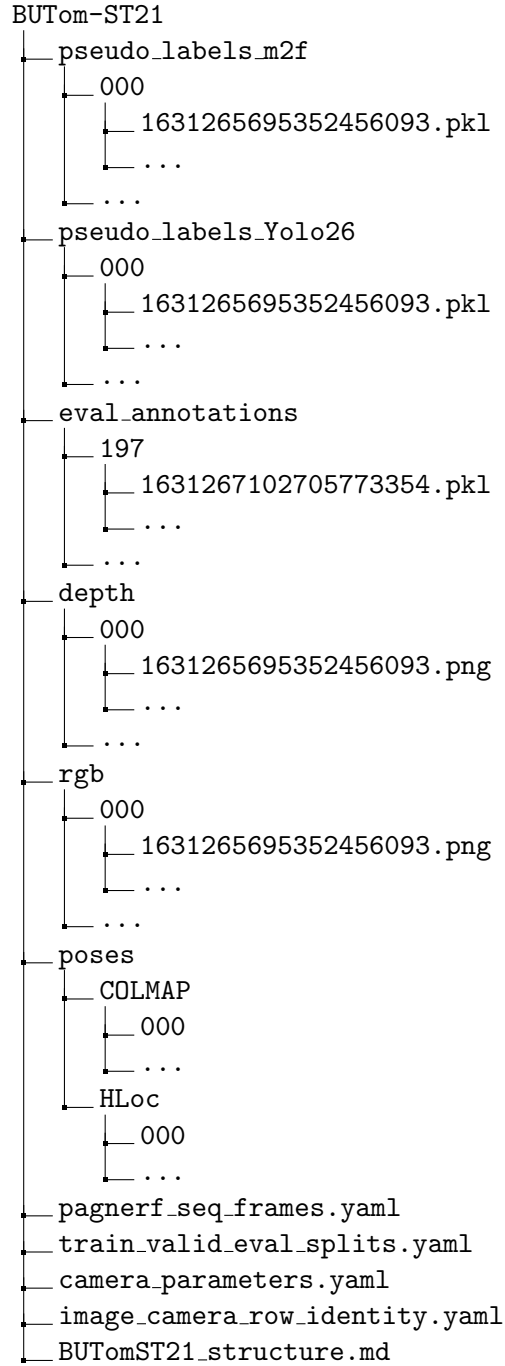

\begin{center}
\begin{minipage}{0.45\textwidth}
\dirtree{%
.1 BUTom-ST21.
.2 pseudo\_labels\_m2f.
.3 000.
.4 1631265695352456093.pkl.
.4 ....
.3 ....
.2 pseudo\_labels\_Yolo26.
.3 000.
.4 1631265695352456093.pkl.
.4 ....
.3 ....
.2 eval\_annotations.
.3 197.
.4 1631267102705773354.pkl.
.4 ....
.3 ....
.2 depth.
.3 000.
.4 1631265695352456093.png.
.4 ....
.3 ....
.2 rgb.
.3 000.
.4 1631265695352456093.png.
.4 ....
.3 ....
.2 poses.
.3 COLMAP.
.4 000.
.4 ....
.3 HLoc.
.4 000.
.4 ....
.2 pagnerf\_seq\_frames.yaml.
.2 train\_valid\_eval\_splits.yaml.
.2 camera\_parameters.yaml.
.2 image\_camera\_row\_identity.yaml.
.2 BUTomST21\_structure.md.
}
\end{minipage}
\end{center}
\caption{BUTom-ST21 data structure.}
\label{fig:butomst21struct}
\end{figure}

For consistency, the spatial-temporal dataset, BUTom-ST21 contains a similar file structure.
The Mask2Former and Yolo26 annotation pickles are supplied in their own directory for the training and validation subsets.
For the evaluation labels, we supply them in a separate directory as this ensures that researchers can select between the appropriate pseudo-labels but still utilize the hand-corrected evaluation subset.
A visualization of the file structure is provided in Figure~\ref{fig:butomst21struct}.

In the \textit{train\_valid\_eval\_splits.yaml} we supply the sequences that belong to each subsection.
The \textit{camera\_parameters.yaml} file provides the intrinsic parameters for both cameras used in the dataset. 
The \textit{image\_camera\_row\_identity.yaml} is similar to that of BUTom21 and supplies information about what camera and on which date the sequences where captured. 
This YAML file also specifies which BUTom21 image is located at the center of its associated sequence.
Finally, the \textit{BUTomST21\_structure.md} file provides indepth information about each file and its structure.

Each BUTom-ST21 sequence consists of 83 frames, where the center 63 have been annotated. 
All pickle annotations (Mask2Former/Yolo26 pseudo-labels and the hand-corrected evaluation labels) are supplied for these 63 central frames. 
We release the complete 83 frame RGB, depth, and pose data (the full input for PAg-NeRF).
This enables researchers to re-run PAg-NeRF or apply an alternative method.
The \textit{annotated\_seq\_frame\_list.yaml} file lists the 63 annotated frames for each sequence, while \textit{pagnerf\_seq\_list.yaml} displays the full 83-frame input sequences fed to PAg-NeRF.

The RGB and depth images are separated based on their sequence number and the various pickle annotations are similarly arranged.
The pseudo-labels are split into Mask2Former or Yolo26 based labels for comparison of the different approaches.
Each of these pseudo-label directories are those trained on BUTom21 using the various image-based semantic segmentation approaches, refined with PAg-NeRF, and post-processed by us.
The hand-corrected evaluation labels are stored in the \textit{eval\_annotations} directory.
These are only supplied for the 22 evaluation sequences.

All annotations are supplied as Python-based pickle files (.pkl extensions).
Each pickle file stores the annotation of a single frame as a dictionary keyed by the tracklet identifier (\textit{track\_id}) of every tomato instance in that frame.
The \textit{track\_id} provides temporal consistency: the same physical tomato maintains the same \textit{track\_id} throughout a sequence, enabling instances to be tracked over time by matching keys between consecutive pickles.
This makes BUTom-ST21 suitable for both VIS and MOT. 
These identifiers are guaranteed to be consistent in the hand-corrected evaluation subset and weakly consistent in the pseudo-labeled training and validation subsets.
Each \textit{track\_id} maps to a dictionary containing four fields: the bounding box in Coco format ([x, y, w, h]), the area (mask pixel count), the semantic label (ripeness subclass), and the instance mask (full-resolution binary mask).

\section{Technical Validation}

To evaluate the viability of our two novel datasets we analyze their performance using state-of-the-art techniques for a variety of tasks.
Our primary tasks for the still image dataset include: instance-based semantic segmentation, bounding box regression, and cross-domain anomaly segmentation.
For BUTom-ST21 we evaluate the performance of our pseudo-labels for instance-based semantic segmentation, and bounding box regression; both evaluated on BUTom21 for completeness of analysis.
Finally, we evaluate video instance segmentation (VIS) and multi-object tracking (MOT) as these are the main focus of the dataset. 

\subsection*{Analysis on BUTom21} 
To analyze the performance of our novel still image BUTom21 tomato dataset, we first evaluate the performance for instance-based semantic segmentation.
Table~\ref{tab:butom21_segAP} outlines these results.
Here we evaluate Mask2Former and Yolo26 models, and train four models per approach: single-class, multi-class, depth filtered single-class and depth filtered multi-class (where depth filtering is completed in a similar manner to that described above).
For Mask2Former we train for 650 epochs maximum, using a total batch size of 32 across 8 NVIDIA A100 GPUs. 
Yolo26 is trained for 100 epochs maximum at all internal configuration parameters and we use a batch size of 16 split evenly over 8 NVIDIA RTX A6000 GPUs. 
For both approaches, the best model is selected based on the validation subset.

\begin{table}[hb]
    \caption{Results for BUTom21 still image dataset instance-based semantic segmentation results trained on the BUTom21 training set. Evaluated using various average precision metrics. The ticks in depth-filtered indicate whether the model used depth-filtered annotation or not, and the multi-class option indicates whether multi-class (tick) or single-class (cross) was used during training.}
    \label{tab:butom21_segAP}
    \resizebox{\textwidth}{!}{
    \begin{tabular}{|l|c|c|c|c|c|c|c|c|c|c|}
        \hline
        Technique & Depth Filtered & Multi-Class & mAP & AP50 & AP75 & AP$_{Tiny}$ & AP$_{Small}$ & AP$_{AllSmall}$ & AP$_{Medium}$ & AP$_{Large}$ \\
        \hline
        Mask2Former & \cmark & \cmark & 29.0 & 41.6 & 32.9 & 1.1 & 28.5 & 25.4 & 56.9 & - \\
        Mask2Former & \xmark & \cmark & 24.6 & 39.4 & 26.4 & 4.6 & 28.2 & 22.8 & 59.0 & - \\
        Mask2Former & \cmark & \xmark & 49.0 & 72.3 & 54.1 & 0.9 & 50.1 & 46.3 & 87.1 & - \\
        Mask2Former & \xmark & \xmark & 40.1 & 64.1 & 43.1 & 30. & 47.8 & 38.8 & 84.8 & - \\
        Yolo26 & \cmark & \cmark & 30.1 & 48.7 & 32.7 & 2.9 & 33.4 & 30.3 & 41.6 & - \\
        Yolo26 & \xmark & \cmark & 28.0 & 49.2 & 28.0 & 9.6 & 32.5 & 27.4 & 49.3 & - \\
        Yolo26 & \cmark & \xmark & 44.8 & 74.4 & 48.4 & 4.7 & 46.8 & 43.6 & 71.5 & - \\
        Yolo26 & \xmark & \xmark & 40.2 & 72.8 & 39.4 & 13.1 & 46.2 & 39.5 & 75.7 & - \\
        \hline
    \end{tabular}
    }
\end{table}

Table~\ref{tab:butom21_segAP} outlines the instance-based semantic segmentation of our approaches.
We evaluate for mean average precision (mAP), average precision at 50\% and 75\% and the mAP of the different object sizes.
In our results we also provide an overall summary on the performance of ``Tiny'' and ``Small'' objects by grouping them together in ``AllSmall''.
For multi-class evaluations, we find that Yolo26 outperforms Mask2Former, showcasing its ability to better distinguish between subclasses.
For both single-class evaluations we either see commensurate performance (not depth filtered) or an improvement when using Mask2Former for depth filtered.
This indicates that for depth filtered approaches Mask2Former is generally able to perceive tomatoes in the foreground better than Yolo26.
However, if you wish to do multi-class segmentation, then, on this evidence, Yolo26 is a better solution.

For bounding box regression, see Table~\ref{tab:butom21_bboxAP}, we see a shift in performance.
Yolo26 is considerably better at detecting objects in the scene.
We attribute this to the way Yolo26 functions, prioritizing detection over segmentation.
These results show, supported by further experiments, that Mask2Former should be chosen for better segmented masks; conversely, Yolo26 is the obvious choice for better detections.

\begin{table}[hb]
    \caption{Results for BUTom21 still image dataset bounding box results trained on the BUTom21 training set. Evaluated using various average precision metrics. The ticks in depth-filtered indicate whether the model used depth-filtered annotation or not, and the multi-class option indicates whether multi-class (tick) or single-class (cross) was used during training.}
    \label{tab:butom21_bboxAP}
    \resizebox{\textwidth}{!}{
    \begin{tabular}{|l|c|c|c|c|c|c|c|c|c|c|}
        \hline
        Technique & Depth Filtered & Multi-Class & mAP & AP50 & AP75 & AP$_{Tiny}$ & AP$_{Small}$ & AP$_{AllSmall}$ & AP$_{Medium}$ & AP$_{Large}$ \\
        \hline
        Mask2Former & \cmark & \cmark & 29.2 & 39.7 & 33.0 & 0.6 & 31.3 & 27.9 & 46.2 & - \\
        Mask2Former & \xmark & \cmark & 24.4 & 36.6 & 26.9 & 5.9 & 29.0 & 23.7 & 46.2 & - \\
        Mask2Former & \cmark & \xmark & 50.9 & 70.7 & 56.6 & 1.9 & 53.5 & 49.5 & 83.9 & - \\
        Mask2Former & \xmark & \xmark & 40.7 & 61.5 & 45.1 & 4.5 & 49.2 & 40.0 & 75.6 & - \\
        Yolo26 & \cmark & \cmark & 36.4 & 49.3 & 41.1 & 4.3 & 40.8 & 37.1 & 46.7 & - \\
        Yolo26 & \xmark & \cmark & 34.7 & 50.5 & 38.9 & 14.4 & 40.3 & 34.4 & 52.8 & - \\
        Yolo26 & \cmark & \xmark & 55.7 & 75.8 & 63.3 & 6.6 & 58.7 & 54.6 & 80.7 & - \\
        Yolo26 & \xmark & \xmark & 51.7 & 75.0 & 57.6 & 19.3 & 59.3 & 51.1 & 83.3 & - \\
        \hline
    \end{tabular}
    }
\end{table}

Our next evaluation is based on anomaly semantic segmentation.
To analyze our results, we utilize a broad set of metrics similar to~\cite{chan2021segmentmeifyoucan}, and follow their evaluation protocols.
For binary classification, we outline the area under the receiver operating characteristic curve (AUROC) metric.
We also include two ranking metrics: area under the precision-recall curve (AUPR), false positive rate at 95\% true positive rate (FPR95).
Among localization metrics, we employ the (foreground) segment-wise intersection over union (sIoU), positive prediction value (PPV), and mean F1 score (meanF1).

For our results assessing the performance of uncertainty estimation methods to cross-crop (out of distribution) OoD objects, we employ various state-of-the-art techniques.
These results are displayed in Table~\ref{tab:ood_results}.
Our two baselines, Mask2Former (\textit{ADE20k Weights}) and Mask2Former (\textit{BUP20 Weights}) are displayed at the top of the table.
For the model trained on BUP20 weights, we fine-tune a separate Mask2Former model to generate single-class (tomatoes) semantic segmentation masks.
In our work, we consider tomatoes to be the anomalous objects.
We benchmark four uncertainty qualification approaches:
Snapshot ensemble~\cite{huang2017snapshot}, Monte Carlo dropout~\cite{gal2016dropout}, single-trajectory LoRA ensemble as representatives of Bayesian approaches, and deep deterministic uncertainty~\cite{mukhoti2023deep} as a representative single-shot deterministic approach.
Then, inspired by~\cite{wald2021calibration} \footnote{Unlike \cite{wald2021calibration}, which fits calibration methods on samples from two domains to improve cross-domain generalization, we fit and evaluate the calibration methods using only the target-domain (tomato) samples.}, we benchmark six post-hoc calibration techniques:
temperature scaling~\cite{guo2017calibration}, logistic scaling~\cite{platt1999probabilistic}, Dirichlet scaling~\cite{kull2019beyond}, Meta-cal~\cite{ma2021meta}, local temperature scaling~\cite{ding2021local}, and selective scaling methods drawn from \cite{wang2023calibrating}.
All ten approaches use Mask2Former as their base segmentation approach.

\begin{table}
    \caption{Binary Anomaly segmentation detection results for cross-domain transfer from BUP20 (sweet pepper) to BUTom21 (tomato).. Scorer: Ent.\ = predictive entropy; MI = mutual information; Den = Density. $\uparrow$: higher is better; $\downarrow$: lower is better. All results report mean\,$\pm$\,std over 5 seeds. Best result per metric in \textbf{bold}. All results displayed for the anomaly detection segmentation include Mask2Former in their pipeline, we remove this here in their ``Method'' column to increase the size of the table.}
    \vspace{2mm}
    \label{tab:ood_results}
    \resizebox{\textwidth}{!}{
    \begin{tabular}{|l|c|c|c|c|c|c|c|}
        \hline
        Method & Scorer & AUROC $\uparrow$ & AUPR $\uparrow$ & FPR95 $\downarrow$ & sIoU $\uparrow$ & PPV $\uparrow$ & MeanF1 $\uparrow$ \\
        \hline
        \hline
        M2F (\textit{ADE20k Weights}) & Ent. & 0.2949 & 0.0236 & 0.9833 & 0.0477 & 0.0060 & 0.0071 \\
        \hline
        M2F (\textit{BUP20 Weights}) & Ent.& 0.9589\,$\pm$\,0.0058 
        & 0.4666\,$\pm$\,0.0557 
        & 0.1286\,$\pm$\,0.0544 
        & 0.5204\,$\pm$\,0.0255 
        & 0.2036\,$\pm$\,0.0301 
        & 0.2601\,$\pm$\,0.0285 \\
        \hline
        \hline
        Snapshot Ensemble & Ent. & 0.9734\,$\pm$\,0.0053 & 0.5826\,$\pm$\,0.0310 & 0.0675\,$\pm$\,0.0165 & \textbf{0.5745\,$\pm$\,0.0096} & 0.2514\,$\pm$\,0.0117 & 0.3157\,$\pm$\,0.0119 \\
        \hline
        Snapshot Ensemble & MI. & 0.9709\,$\pm$\,0.0065 & \textbf{0.6179\,$\pm$\,0.0575} & 0.0801\,$\pm$\,0.0244 & 0.5715\,$\pm$\,0.0097 & 0.1591\,$\pm$\,0.0173 & 0.2312\,$\pm$\,0.0199 \\
        \hline
        Monte Carlo Dropout & Ent. & 0.9474\,$\pm$\,0.0185 & 0.4525\,$\pm$\,0.0929 & 0.2201\,$\pm$\,0.1774 & 0.5349\,$\pm$\,0.0364 & 0.1823\,$\pm$\,0.0596 & 0.2420\,$\pm$\,0.0681 \\
        \hline
        Monte Carlo Dropout & MI. & 0.8309\,$\pm$\,0.1032 & 0.2950\,$\pm$\,0.1155 & 0.6763\,$\pm$\,0.1729 & 0.5116\,$\pm$\,0.0675 & 0.1167\,$\pm$\,0.0317 & 0.1504\,$\pm$\,0.0446 \\
        \hline
        Deep Determinsitc Approach & Den. & 0.9365\,$\pm$\,0.0343 
        & 0.5073\,$\pm$\,0.0609 
        & 0.3020\,$\pm$\,0.2299 
        & 0.5484\,$\pm$\,0.0088 
        & 0.1354\,$\pm$\,0.0146 
        & 0.1861\,$\pm$\,0.0127\\
        \hline
        Single Trajectory LoRA Ens  & Ent. & \textbf{0.9752\,$\pm$\,0.0040} & 0.5629\,$\pm$\,0.0374 & \textbf{0.0445\,$\pm$\,0.0015} & 0.5693\,$\pm$\,0.0055 & \textbf{0.2946\,$\pm$\,0.0154} & \textbf{0.3568\,$\pm$\,0.0111} \\
        \hline
        Single Trajectory LoRA Ens & MI. & 0.9737\,$\pm$\,0.0045 & 0.5995\,$\pm$\,0.0707 & 0.0486\,$\pm$\,0.0032 & 0.5704\,$\pm$\,0.0154 & 0.2039\,$\pm$\,0.0205 & 0.2780\,$\pm$\,0.0200 \\
        \hline
        Temperature Scaling & Ent. & 0.8750\,$\pm$\,0.0542 & 0.2514\,$\pm$\,0.0345 & 0.5717\,$\pm$\,0.3430 & 0.4213\,$\pm$\,0.0116 & 0.0778\,$\pm$\,0.0290 & 0.1135\,$\pm$\,0.0312 \\
        \hline
        Logistic Calibration & Ent. & 0.9560\,$\pm$\,0.0053 & 0.3885\,$\pm$\,0.0116 & 0.1219\,$\pm$\,0.0513 & 0.5350\,$\pm$\,0.0082 & 0.1913\,$\pm$\,0.0145 & 0.2553\,$\pm$\,0.0125 \\
        \hline
        Dirichlet Calibration & Ent. & 0.9561\,$\pm$\,0.0053 & 0.3868\,$\pm$\,0.0113 & 0.1219\,$\pm$\,0.0513 & 0.5398\,$\pm$\,0.0081 & 0.1975\,$\pm$\,0.0140 & 0.2618\,$\pm$\,0.0126 \\
        \hline
        Local Temperature Scaling & Ent. & 0.8674\,$\pm$\,0.0406 & 0.2315\,$\pm$\,0.0027 & 0.5107\,$\pm$\,0.2243 & 0.4249\,$\pm$\,0.0100 & 0.0875\,$\pm$\,0.0112 & 0.1283\,$\pm$\,0.0125 \\
        \hline
        Meta-Cal & Ent. & 0.8751\,$\pm$\,0.0544 & 0.2548\,$\pm$\,0.0377 & 0.5717\,$\pm$\,0.3431 & 0.4213\,$\pm$\,0.0116 & 0.0778\,$\pm$\,0.0290 & 0.1135\,$\pm$\,0.0312 \\
        \hline
        Selective Scaling & Ent. & 0.8807\,$\pm$\,0.0608 & 0.2950\,$\pm$\,0.0927 & 0.5555\,$\pm$\,0.3628 & 0.4453\,$\pm$\,0.0542 & 0.1046\,$\pm$\,0.0585 & 0.1463\,$\pm$\,0.0718 \\
        \hline
    \end{tabular}
    }
\end{table}

From these results, we show that semantic segmentation for OoD can be improved over the baseline Mask2Former \textit{BUP20 weights} approach, but not consistently.
In fact, from these results we show that Mask2Former when trained on BUP20 and cross-evaluated out-of-domain on BUTom21 actually performs admirably.
We believe that this work shows the potential of the dataset to be used for future anomaly detection.

\subsection*{Analysis on BUTom-ST21} 

Our first BUTom-ST21 analysis is completed to show the overall performance of the dataset as a still image instance-based semantic segmentation and then detection dataset.
First, the sequences are converted to a still image format (only mask or bounding boxes are considered).
We once again train Mask2Former and Yolo26 based on this considerably larger dataset, where we use the same hyper-parameter as the BUTom21 still image evaluation, however, we only train for 50 epochs for both approaches.
The best model for all is chosen based on the spatial-temporal validation set, and final evaluations are performed on the BUTom21 (still image) evaluation subset.
Due to the manner in which BUTom-ST21 is trained, we only consider the depth filtered BUTom21 evaluation set as we cannot train a BUTom-ST21 not depth filtered model.
This allows us to directly compare BUTom21 to BUTom-ST21, consistent with our prior work~\cite{2025bupst20}.

Ultimately, the challenges associated with psuedo-labels are shown in Tables~\ref{tab:butomst21_segAP}-\ref{tab:butomst21_bboxAP}, when compared to Tables~\ref{tab:butom21_segAP}-\ref{tab:butom21_bboxAP}.
There is between 5\% and 24\% drop in performance from BUTom21 to BUTom-ST21 for segmentation-based mAP.
For detection (bounding boxes) there is a degradation in performance from 6\% to 13\%.
These challenging results are caused by the poor segmentation outputs (see Figure~\ref{fig:butomst21_ex}) from both models.
Overall, this makes BUTom-ST21 an interesting dataset for instance-based semantic segmenation, and perhaps uncertainty estimation could assist in assigning better segmentation masks. 

\begin{table}[hb]
    \caption{Results for BUTom-ST21 still image dataset segmentation AP results trained on the BUTom-ST21 training set. Evaluated using various average precision metrics on the BUTom21 dataset. Validation parameters are also tuned on the BUTom-ST21 dataset. For the multi-class column, ticks indicate multi-class and crosses indicate single class.}
    \label{tab:butomst21_segAP}
    \vspace{2mm}
    \resizebox{\textwidth}{!}{
    \begin{tabular}{|l|c|c|c|c|c|c|c|c|c|}
        \hline
        Technique & Multi-Class & mAP & AP50 & AP75 & AP$_{Tiny}$ & AP$_{Small}$ & AP$_{AllSmall}$ & AP$_{Medium}$ & AP$_{Large}$ \\
        \hline
        Mask2Former & \cmark & 27.4 & 42.2 & 30.1 & 0.0 & 26.3 & 23.7 & 58.3 & - \\
        Mask2Former & \xmark & 42.4 & 69.1 & 44.9 & 0.1 & 43.1 & 39.8 & 84.2 & - \\
        Yolo26 & \cmark & 22.9 & 39.9 & 23.9 & 0.3 & 28.1 & 25.0 & 36.6 & - \\
        Yolo26 & \xmark & 37.5 & 69.8 & 35.6 & 0.6 & 39.3 & 36.1 & 70.7 & - \\
        \hline
    \end{tabular}
    }
\end{table}

\begin{table}[hb]
    \caption{Results for BUTom-ST21 still image dataset bounding box AP results trained on the BUTom-ST21 training set. Evaluated using various average precision metrics on the BUTom21 dataset. Validation parameters are also tuned on the BUTom-ST21 dataset. For the multi-class column, ticks indicate multi-class and crosses indicate single class.}
    \label{tab:butomst21_bboxAP}
    \vspace{2mm}
    \resizebox{\textwidth}{!}{
    \begin{tabular}{|l|c|c|c|c|c|c|c|c|c|}
        \hline
        Technique & Multi-Class & mAP & AP50 & AP75 & AP$_{Tiny}$ & AP$_{Small}$ & AP$_{AllSmall}$ & AP$_{Medium}$ & AP$_{Large}$ \\
        \hline
        Mask2Former & \cmark & 26.4 & 40.1 & 28.8 & 0.0 & 27.0 & 24.4 & 50.7 & - \\
        Mask2Former & \xmark & 42.5 & 66.6 & 45.8 & 0.3 & 43.5 & 40.2 & 81.0 & - \\
        Yolo26 & \cmark & 28.5 & 40.5 & 34.2 & 0.4 & 35.6 & 31.9 & 40.3 & - \\
        Yolo26 & \xmark & 48.5 & 72.5 & 56.3 & 1.4 & 51.5 & 47.3 & 77.4 & - \\
        \hline
    \end{tabular}
    }
\end{table}

Our next experiments use BUTom-ST21 for video instance segmentation (VIS).
The VIS task requires evaluating both temporal consistency (tracklet IDs) and segmentation quality. 
To achieve this, we employ the state-of-the-art technique CTVIS~\cite{ying2023ctvis} fine-tuning it with pretrained Coco weights on the BUTom-ST21 training set and validate it on the validation set. 
We use the Swin-L backbone with default CTVIS configuration, except for the learning rate: 0.0001 for multi-class and 0.00005 for single-class. The model is trained for a maximum of 350 epochs when using Mask2Former pseudo-labels and 600 epochs for the Yolo26 variant. 
We split a batch of 8 across 8 Nvidia A100 GPUs, with an image resolution downsampled to $960\times{540}$.
Results are provided in Table~\ref{tab:butomst21_videoAP} where we report mAP, AP50 and AP75 across object sizes. Unlike the still-image AP variants, we use video-based average precision instead of image-based average precision.

In all cases, the Mask2Former labels provide better segmentation results across their respective sequences (video AP).
This is consistent with the results in Table~\ref{tab:butomst21_segAP} where Mask2Former produced better segmentation quality.
Additionally, consistent with expectations, the single-class approach performs better than the multi-class approach.
We attribute this to the accuracy of the model in classifying the subclass information.
Of particular interest to the research community, performance for ``tiny'' objects is very low, and we achieve a performance of 0.0 regardless of the model.
This once again highlights the challenge and noise in the pseudo-labels, where ``tiny'' objects become unrecoverable.

\begin{table}
    \caption{Video instance segmentation results for CTVIS. It's trained on the BUTom-ST21 training set, and evaluated on hand-corrected evaluation set. The multi-class column indicates whether we used multi-class (tick) models, or single-class (cross).}
    \label{tab:butomst21_videoAP}
    \resizebox{\textwidth}{!}{
    \begin{tabular}{|l|c|c|c|c|c|c|c|c|c|c|}
        \hline
        Technique & Pseudo-labels & Multi-Class & AP & AP50 & AP75 & AP$_{Tiny}$ & AP$_{Small}$ & AP$_{AllSmall}$ & AP$_{Medium}$ & AP$_{Large}$ \\
        \hline
        CTVIS & Mask2Former & \cmark & 27.5 & 43.1 & 30.7 & 0.0 & 15.3 & 15.2 & 43.4 & - \\
        CTVIS & Mask2Former & \xmark & 45.6 & 75.5 & 48.6 & 0.0 & 34.9 & 34.7 & 78.3 & - \\
        CTVIS & Yolo26 & \cmark & 25.4 & 41.7 & 28.3 & 0.0 & 16.1 & 16.0 & 39.6 & - \\
        CTVIS & Yolo26 & \xmark & 46.0 & 77.7 & 48.4 & 0.0 & 35.6 & 35.4 & 77.8 & - \\
        \hline
    \end{tabular}
    }
\end{table}

Our final benchmarking approach on BUTom-ST21 is for multi-object tracking (MOT).
As with all benchmarks we do not perform an exhaustive hyper-parameter sweep; instead, we use the results from BUP-ST21 as our starting point.
We acknowledge that a more thorough hyper-parameter sweep could improve performance across all methods by a few absolute points. 
We compare learned models (CTVIS, PAg-NeRF) against methods without an inherent learning pipeline (ByteTrack~\cite{Bytetrack2022}, Halstead~\cite{halstead2018}, and OCSort~\cite{cao2022observation}).
For the latter approaches, we use the depth filtered outputs from the BUTom21 model and run inference on all sequences in the evaluation subset.

To evaluate the performance of MOT on this dataset, we explore higher-order tracking accuracy (HOTA) for tracking association and object detection; multi-object tracking accuracy (MOTA) to evaluate false positives, false negatives, and ID switching; and IDF1 for frame-by-frame accuracy. We also show results for ID switching, false positives, and false negatives.

\begin{table}
    \caption{Results of multi-object tracking using the multi-class models - for non-learned models, the segmentation masks are post-processed with depth filtering. The multi-class column indicates whether we used multi-class (tick) models, or single-class (cross).}
    \label{tab:motmc}
    \resizebox{\textwidth}{!}{
    \begin{tabular}{|l|c|c|c|c|c|c|c|}
    	\hline
    	Technique & Mask2Former & HOTA $\uparrow$ & MOTA $\uparrow$ & IDF1 $\uparrow$ & ID switch $\downarrow$ & FP $\downarrow$ & FN $\downarrow$ \\
    	\hline
    	ByteTrack - IoU & \cmark & 69.2 & 34.1 & 65.2 &1039 & 16582 & 11292 \\
    	ByteTrack - IoU & \xmark & \textbf{79.9} & 57.7 & 76.8 & \textbf{401} & 8997 & \textbf{9176} \\
    	Halstead - IoU & \cmark & 52.2 & 30.9 & 44.9 &3108 & 15048 & 12179 \\
    	Halstead - IoU & \xmark & 74.1 & \textbf{59.0} & 69.6 &925 & \textbf{7441} & 9641 \\
        OCSort & \cmark & 65.5 & 35.9 & 62.6 & 2156 & 14620 & 11337 \\
        OCSort & \xmark & 78.7 & 58.7 & \textbf{76.2} & 650 & 8523 & 9223 \\
    	\hline
    	PAg-NeRF       & \cmark & 66.1 & 48.6 & 57.4 & \textbf{31} & \textbf{6726} & 15786 \\
    	PAg-NeRF       & \xmark & 65.1 & 48.5 & 57.8 & 69 & 7842 & 14704 \\
    	CTVIS          & \cmark & 83.0 & 61.1 & 77.6 & 399 & 7385 & 9300 \\
    	CTVIS          & \xmark & \textbf{83.8} & \textbf{64.1} & \textbf{80.0} & 318 & 7872 & \textbf{7577} \\
    	\hline
    \end{tabular}
    }
\end{table}

\begin{table}
    \caption{Results of multi-object tracking using the single-class models. The multi-class column indicates whether we used multi-class (tick) models, or single-class (cross).}
    \label{tab:motsc}
    \resizebox{\textwidth}{!}{
    \begin{tabular}{|l|c|c|c|c|c|c|c|}
    	\hline
    	Technique & Mask2Former & HOTA $\uparrow$ & MOTA $\uparrow$ & IDF1 $\uparrow$ & ID switch $\downarrow$ & FP $\downarrow$ & FN $\downarrow$ \\
    	\hline
    	ByteTrack - IoU & \cmark & 77.8 & 54.9 & 74.2 & 435 & 9053 & 10304 \\
    	ByteTrack - IoU & \xmark & 79.1 & 57.8 & \textbf{76.8} & \textbf{310} & 8547 & 9651 \\
    	Halstead - IoU & \cmark & 71.3 & 57.0 & 67.0 & 1022 & \textbf{7149} & 10682 \\
    	Halstead - IoU & \xmark & \textbf{82.0} & 58.3 & 67.7 & 1024 & 7203 & \textbf{9641} \\
        OCSort & \cmark & 76.8 & 57.9 & 75.4 & 661 & 7156 & 10677 \\
        OCSort & \xmark & 78.1 & \textbf{58.8} & 76.5 & 596 & 7822 & 9663 \\
    	\hline
    	CTVIS          & \cmark & \textbf{84.6} & 61.6 & 78.8 & \textbf{234} & 8060 & 8552 \\
    	CTVIS          & \xmark & 84.5 & \textbf{66.2} & \textbf{81.1} & 272 & \textbf{7638} & \textbf{6914}\\
    	\hline
    \end{tabular}
    }
\end{table}

In Table~\ref{tab:motmc}, we see the best MOT performance in the learned approaches.
These approaches benefit from learning all information in their training paradigm.
CTVIS using Yolo26 labels is able to achieve the best HOTA, MOTA, IDF1, and false negatives, while PAg-NeRF using Mask2Former outputs achieves the best ID switches and false positives.
Among standard (unlearned) approaches, we see a variation in best performance, except for the fact that in all cases Yolo26 performs the best.
We performed a further analysis of this anomaly and found that Mask2Former produced considerable overlapping outputs creating additional challenges in an already difficult task.
From our analysis, we found that by performing mask-based non-maximal supression (NMS) we were able to improve the ByteTrack Mask2Former performance to 76.6 (HOTA), 55.3 (MOTA), 73.8 (IDF1), 471 (ID switches), 7627 (false positives), and 11533 (false negatives).
This is a considerable boost in performance and shows one of the limitations associated with instance-based semantic segmentation in Mask2Former.

For single-class evaluation (Table~\ref{tab:motsc}) we exclude PAg-NeRF as it was trained only on multi-class outputs.
Once again, we see the learned re-identification strength of CTVIS outperforms all other approaches, though not by as considerable a margin as that witnessed in Table~\ref{tab:motmc}.
The performance of Mask2Former outputs is much more inline with its Yolo26 counterpart, with Yolo26 matching or exceeding Mask2Former.

Overall, we have shown how our two BUTom* datasets can be used and established a number of benchmark values across multiple computer vision tasks in agriculture.
The challenges of the two datasets are clearly evident, particularly for ``tiny'' objects.

\section{Usage Notes (optional)}

To aid with reproducibility, both BUTom* datasets have three uniquely defined subsets: training, validation, and evaluation.
One of the key differences this paper has over the BUP-ST21 is the number of small and tiny objects.
Table~\ref{tab:bdbuptom21} has a summary of these statistics.
To separate the size of objects we split them into the following groups:
tiny: 0-100 (10$\times$10) pixels;
small: 101-3,025 (55$\times$55) pixels; 
medium: 3,026-27,556 (166$\times$166) pixels; and
large: anything larger than 27,556 pixels.
These object sizes are inspired by the Coco dataset~\cite{lin2014microsoft} which we have refactored based on our image resolution.
This conversion is required as Coco images are of the resolution $640\times{480}$, while ours are $1280\times{720}$.

To calculate the above object sizes we used the ratio of total pixels between the two datasets.
We obtained a ratio of 1:3 as the image sizes were $640\times{480}=307200$ pixels and ours $1280\times{720}=921600$.
Then we calculated object sizes based on this to again have similar total number of pixels $\sqrt{32^2\times{3}} \sim 55^2$ pixels and $\sqrt{96^2\times{3}} \sim 166^2$ pixels.
Finally, we selected $10^2$ as our lower limit to the tiny size because the Mask2Former approach downsamples the feature map 4 times, and at the lowest resolution 100 pixels becomes approximately 3 which we could consider as noise.

\section{Data Availability}

The two novel datasets are available for download by following the commands described in \url{https://github.com/Agricultural-Robotics-Bonn/BUTom21-ST21}.
In this github repository contains the instructions for downloading both BUTom21 and BUTom-ST21.
To ensure data storage viability, the two datasets will be hosted on \textit{BonnData} (\url{https://bonndata.uni-bonn.de/}).
These two datasets can be downloaded independently of each other, allowing researchers to decide which dataset best suits their needs or tasks.
Figure~\ref{fig:butom21struct}-\ref{fig:butomst21struct} outlines the general structure of the datasets and what can be expected in each.
The image format is supplied as png files, both for RGB and depth.
For BUTom21 the annotations are supplied as a Coco format json, and for BUTom-ST21 we supply them as pickles per image.
Each pickle contains the full information for each tracklet ID/object in the current scene, where tracklet IDs are consistent throughout the sequence. 
Also camera pose information per sequence is provided for BUTom-ST21.

\section{Code Availability}

We have supplied minimal working code. 
The code is available here: \url{https://github.com/Agricultural-Robotics-Bonn/BUTom21-ST21} and includes minimalist PyTorch style dataloaders for each of the two BUTom* datasets.
We also supply code to create Yolo26 style image directories and annotations from the Coco format and pickle format variants.
No tracking or evaluation code has been supplied.

\section{Author Contributions}

Halstead contributed to the annotation, code and evaluation, and writing the manuscript. Guclu contributed to data preparation, annotation, code and evaluation, and writing the manuscript. Farag contributed to the experimental analysis and manuscript writing. Pallotta contributed to the annotation and manuscript. Hund contributed to annotation, and general organization. The remaining authors contributed to the writing of the manuscript and funding arrangements. 

\section{Competing Interests}

None.

\section{Acknowledgements (optional)}

The authors gratefully acknowledge the access to the \textit{Marvin} cluster of the University of Bonn. 
We the authors would also like to thank Yan Wang, Sina Raufi, Efe Incir, 
Patrick Zimmer, 
Nelson Pinheiro,
Julian Rosbach,
Rafay Aamir,
Moein Taherkhani,
Sicong Pan, and
Gokul Krishna Gandhi Chenchani
for their assistance in annotating the data.

\section{Funding}

This work has been funded by the Deutsche Forschungsgemeinschaft (DFG, German Research Foundation) for KI-FOR 5351: MC 831/2-1 (459376902), RO 4839/6-1 (498564628), and GA 1927/9-1 (498577300). 

\section{Ethics statement*}


\section{References}


\begin{thebibliography}{9}

\bibitem{2025bupst20}
Guclu, E., Halstead, M., Denman, S., \& McCool, C. (2025). 
``Weakly Labelled Spatial-Temporal Sweet Pepper Data: enabling higher quality detection, segmentation, and tracking''. 
\textit{The International Journal of Robotics Research}. 
doi: 10.1177/02783649251379093.

\bibitem{smitt2022pag}
Smitt, C., Halstead, M., Zimmer, P., Läbe, T., Guclu, E., Stachniss, C., \& McCool, C. (2023). 
``PAg-NeRF: Towards fast and efficient end-to-end panoptic 3D representations for agricultural robotics''. 
\textit{IEEE Robotics and Automation Letters}. 
9(1), 907-914.

\bibitem{smittPathobot2021}
Smitt, C., Halstead, M., Zaenker, T., Bennewitz, M., \& McCool, C. (2021). ``Pathobot: A robot for glasshouse crop phenotyping and intervention''. 
\textit{IEEE International Conference on Robotics and Automation}. 
pp. 2324-2330.

\bibitem{halsteadbup19}
Halstead, M., Denman, S., Fookes, C., \& McCool, C. (2020). 
``Fruit detection in the wild: The impact of varying conditions and cultivar''. 
\textit{Digital image computing: techniques and applications}. 
pp. 1-8.

\bibitem{wang2025}
Wang, Y., Fei, Z., Li, R., \& Ying, Y. (2025). ``Learn from foundation model: Fruit detection model without manual annotation''. 
\textit{Pattern Recognition}. 
112799.

\bibitem{hani2020}
Häni, N., Roy, P., \& Isler, V. (2020). 
``MinneApple: a benchmark dataset for apple detection and segmentation''. 
\textit{IEEE Robotics and Automation Letters}. 
5(2), 852-858.

\bibitem{james2024}
James, J. A., Manching, H. K., Mattia, M. R., Bowman, K. D., Hulse-Kemp, A. M., \& Beksi, W. J. (2024). ``Citdet: A benchmark dataset for citrus fruit detection''. 
\textit{IEEE Robotics and Automation Letters}. 
9(12), 10788-10795.

\bibitem{barbosa2024}
Barbosa Junior, M. R., Santos, R. G. D., Sales, L. D. A., \& Oliveira, L. P. D. (2024). ``Advancements in agricultural ground robots for specialty crops: an overview of innovations, challenges, and prospects''. 
\textit{Plants}. 
13(23), 3372.

\bibitem{saizrubio2020}
Saiz-Rubio, V., \& Rovira-Más, F. (2020). ``From smart farming towards agriculture 5.0: A review on crop data management''. 
\textit{Agronomy}. 
10(2), 207.

\bibitem{yepezponce2023}
Yépez-Ponce, D. F., Salcedo, J. V., Rosero-Montalvo, P. D., \& Sanchis, J. (2023). 
``Mobile robotics in smart farming: current trends and applications''. 
\textit{Frontiers in artificial intelligence}. 
6, 1213330.

\bibitem{chengm2f2022}
Cheng, B., Misra, I., Schwing, A. G., Kirillov, A., \& Girdhar, R. (2022). 
``Masked-attention mask transformer for universal image segmentation''. 
\textit{IEEE/CVF conference on computer vision and pattern recognition}.  
pp. 1290-1299.

\bibitem{jocher2026ultralyticsyolo26unifiedrealtime}
Jocher, G., Qiu, J., Liu, M., Lyu, S., Akyon, F. C., \& Kalfaoglu, M. E. (2026). ``Ultralytics YOLO26: Unified Real-Time End-to-End Vision Models''. 
\textit{arXiv preprint arXiv:2606.03748}.

\bibitem{cocoannotator}
Brooks, J. (2019). 
``COCO Annotator''. 
\url{https://github.com/jsbroks/coco-annotator/}.

\bibitem{arad2020development}
Arad, B., Balendonck, J., Barth, R., Ben-Shahar, O., Edan, Y., Hellstr{\"o}m, T., Hemming, J., Kurtser, P., Ringdahl, O., Tielen, T., \& van Tuijl, B. (2020). 
``Development of a sweet pepper harvesting robot''. 
\textit{Journal of Field Robotics}. 
37, 1027–1039.

\bibitem{zhang2026tomato}
Zhang, Y., Struckmeyer, S., Kolb, A., \& Reichardt, S. (2026). 
``Tomato Multi-Angle Multi-Pose Dataset for Fine-Grained Phenotyping''. 
\textit{Scientific Data}. 
13:309.


\bibitem{afonso2020tomato}
Afonso, M., Fonteijn, H., Fiorentin, F. S., Lensink, D., Mooij, M., Faber, N., Polder, G., \& Wehrens, R. (2020). 
``Tomato fruit detection and counting in greenhouses using deep learning''. 
\textit{Frontiers in Plant Science}. 
11:571299.

\bibitem{laborotomato2023}
Laboro AI. (2023). 
``Laboro tomato: Instance segmentation dataset. GitHub Repository''. 
\url{https://github.com}. 
(Accessed on January 5, 2023).

\bibitem{tsironis2020tomatod}
Tsironis, V., Bourou, S., \& Stentoumis, C. (2020). 
``TomatOD: Evaluation of object detection algorithms on a new real-world tomato dataset''. 
\textit{ISPRS - International Archives of the Photogrammetry, Remote Sensing and Spatial Information Sciences}. 

\bibitem{de2022apple}
de Jong, S., Baja, H., Tamminga, K., \& Valente, J. (2022). 
``Apple mots: Detection, segmentation and tracking of homogeneous objects using mots''. 
\textit{IEEE Robotics and Automation Letters}. 
7:11418–11425.

\bibitem{Josegovmp2025}
Jose, A. I., Pan, S., Zaenker, T., Menon, R., Houben, S., \& Bennewitz, M. (2025). 
``Go-vmp: Global optimization for view motion planning in fruit mapping''. 
\textit{International Conference on Intelligent Robots and Systems (IROS)}.  
pp. 1825-1832.


\bibitem{eurostat2025crops}
Eurostat. (2025). ``Agricultural Production - Crops''. 
\url{https://ec.europa.eu/eurostat/statistics-explained/index.php?title=Agricultural_production_-_crops#Fruit}.
Accessed 2026-07-09

\bibitem{cbi2023tomato}
Centre for the Promotion of Imports, Netherlands Ministry of Foreign Affairs. (2023). 
``The European Market Potential for Tomatoes''. 
\url{https://www.cbi.eu/market-information/fresh-fruit-vegetables/tomatoes/market-potential}.
Accessed 2026-07-09

\bibitem{halstead2021agnostic}
Halstead, M., Ahmadi, A., Smitt, C., Schmittmann, O., \& McCool, C. (2021). 
``Crop agnostic monitoring driven by deep learning''. 
\textit{Frontiers in plant science}. 
12, 786702.

\bibitem{CarionSam3}
Carion, N., Gustafson, L., Hu, Y. T., Debnath, S., Hu, R., Suris, D., ... \& Feichtenhofer, C. (2025). 
``Sam 3: Segment anything with concepts''. 
\textit{arXiv preprint arXiv:2511.16719}.

\bibitem{ravi2025sam}
Ravi, N., Gabeur, V., Hu, Y.-T., Hu, R., Ryali, C., Ma, T., Khedr, H., R{\"a}dle, R., Rolland, C., Gustafson, L., et al. (2025). 
``Sam 2: Segment anything in images and videos''. 
\textit{International Conference on Learning Representations}. 
2025:28085–28128.

\bibitem{sarlin2019coarse}
Sarlin, P.-E., Cadena, C., Siegwart, R., \& Dymczyk, M. (2019). 
``From coarse to fine: Robust hierarchical localization at large scale''. 
\textit{Proceedings of the IEEE/CVF Conference on Computer Vision and Pattern Recognition}. 
12716–12725.

\bibitem{schonberger2016structure}
Schonberger, J. L., \& Frahm, J.-M. (2016). 
``Structure-from-motion revisited''. 
\textit{Proceedings of the IEEE Conference on Computer Vision and Pattern Recognition}. 
4104–4113.

\bibitem{detone2018superpoint}lindenberger2023lightglue
DeTone, D., Malisiewicz, T., \& Rabinovich, A. (2018). 
``Superpoint: Self-supervised interest point detection and description''. 
\textit{Proceedings of the IEEE Conference on Computer Vision and Pattern Recognition Workshops}. 
224–236.

\bibitem{lindenberger2023lightglue}
Lindenberger, P., Sarlin, P.-E., \& Pollefeys, M. (2023). 
``Lightglue: Local feature matching at light speed''. 
\textit{Proceedings of the IEEE/CVF International Conference on Computer Vision}. 
17627–17638.

\bibitem{Zhouetal2018}
Zhou, B., Zhao, H., Puig, X., Xiao, T., Fidler, S., Barriuso, A., \& Torralba, A. (2018)/
``Semantic understanding of scenes through ADE20K dataset''.
\textit{International Journal of Computer Vision}.
127, pp. 302--321.

\bibitem{chan2021segmentmeifyoucan}
Chan, R. K. W., Lis, K., Uhlemeyer, S., Blum, H., Honari, S., Siegwart, R., Fua, P., Salzmann, M., \& Rottman, M. (2021).
``SegmentMeIfYouCan: A benchmark for anomaly segmentation''.
\textit{Proceedings of the Neural Information Processing Systems (NeurIPS) Track on Datasets and Benchmarks}.

\bibitem{sodano2024open}
M.~Sodano, F.~Magistri, L.~Nunes, J.~Behley, and C.~Stachniss,
Sodano, M., Magistri, F., Nunes, L., Behley, J., \& Stachniss, C. (2024).
``Open-world semantic segmentation including class similarity''.
\textit{Proceedings of the IEEE/CVF Conference on Computer Vision and Pattern Recognition}.
pp. 3184--3194.

\bibitem{smitt2021pathobot}
Smitt, C., Halstead, M., Zaenker, T., Bennewitz, M., \& McCool, C. (2021).
``Pathobot: A robot for glasshouse crop phenotyping and intervention''.
\textit{IEEE International Conference on Robotics and Automation}. 
pp. 2324--2330.

\bibitem{huang2017snapshot}
Huang, G., Li, Y., Pleiss, P., Liu, Z., Hopcroft, J. E., Weinberger, K. Q. (2017).
``Snapshot Ensembles: Train 1, Get M for Free''.
\textit{International Conference on Learning Representations}.

\bibitem{gal2016dropout}
Gal, Y., \& Ghahramani, Z. (2016).
``Dropout as a Bayesian approximation: representing model uncertainty in deep learning''.
\textit{Proceedings of the 33rd International Conference on Machine Learning}.
pp. 1050--1059.

\bibitem{mukhoti2023deep}
Mukhoti, J., Kirsch, A., van Amersfoort, J., Torr, P. H. S., Gal, Y. (2023).
``Deep deterministic uncertainty: A new simple baseline''.
\textit{Proceedings of the IEEE/CVF Conference on Computer Vision and Pattern Recognition (CVPR)}.
pp. 24384--24394.

\bibitem{deng2009imagenet}
Deng, J., Dong, W., Socher, R., Li, L.-J., Li, K., \& Fei-Fei, L. (2009). 
``Imagenet: A large-scale hierarchical image database''. 
\textit{IEEE Conference on Computer Vision and Pattern Recognition}. 
pp. 248–255.

\bibitem{lin2014microsoft}
Lin, T. Y., Maire, M., Belongie, S., Hays, J., Perona, P., Ramanan, D., Dollár, P., \& Zitnick, C. L. (2014). 
``Microsoft coco: Common objects in context''. 
\textit{In European conference on computer vision}. 
pp. 740–755.

\bibitem{liu2021swin}
Liu, Z., Lin, Y., Cao, Y., Hu, H., Wei, Y., Zhang, Z., Lin, S., \& Guo, B. (2021). 
``Swin transformer: Hierarchical vision transformer using shifted windows''. 
\textit{Proceedings of the IEEE/CVF International Conference on Computer Vision}. 
pp. 10012–10022.

\bibitem{milan2016mot16}
Milan, A., Leal-Taix{\'e}, L., Reid, I., Roth, S., \& Schindler, K. (2016). 
``MOT16: A benchmark for multi-object tracking''. 
\textit{arXiv preprint arXiv:1603.00831}.

\bibitem{ying2023ctvis}
Ying, K., Zhong, Q., Mao, W., Wang, Z., Chen, H., Wu, L. Y., Liu, Y., Fan, C., Zhuge, Y., \& Shen, C. (2023). 
``Ctvis: Consistent training for online video instance segmentation''. 
\textit{Proceedings of the IEEE/CVF International Conference on Computer Vision}. 
pp. 899–908.

\bibitem{wald2021calibration}
Wald, Y., Feder, A., Greenfield, D., Shalit, U. (2021).
``On calibration and out-of-domain generalization''.
\textit{Advances in Neural Information Processing Systems}. 
vol. 34, pp. 2215-2227.

\bibitem{wang2023calibrating}
Wang, D., Gong, B., \& Wang, L. (2023).
``On calibrating semantic segmentation models: Analyses and an algorithm,''
\textit{Proceedings of the IEEE/CVF Conference on Computer Vision and Pattern Recognition}. 
pp. 23652-23662.

\bibitem{guo2017calibration}
Guo, C., Pleiss, G., Sun, Y., \& Weinberger, K. Q. (2017).
``On calibration of modern neural networks''.
\textit{Proceedings of the 34th International Conference on Machine Learning}. 
pp. 1321-1330.

\bibitem{platt1999probabilistic}
Platt, J. (1999).
``Probabilistic outputs for support vector machines and comparisons to regularized likelihood methods''.
\textit{Advances in Large Margin Classifiers}. 
Vol. 10, pp. 61-74.

\bibitem{kull2019beyond}
Kull, M., Perello-Nieto, M., K{\"a}ngsepp, M., Filho, T. S., Song, H., \& Flach, P. (2019).
``Beyond temperature scaling: Obtaining well-calibrated multiclass probabilities with Dirichlet calibration''.
\textit{Advances in Neural Information Processing Systems}.
Vol. 32.

\bibitem{ma2021meta}
X.~Ma and M.~B. Blaschko,
MA, X., \& Blaschko, M. B. (2021).
``Meta-cal: Well-controlled post-hoc calibration by ranking''.
\textit{International Conference on Machine Learning}.
pp. 7235-7245.

\bibitem{ding2021local}
Ding, Z., Han, X., Liu, P., \& Niethammer, M. (2021).
``Local Temperature Scaling for Probability Calibration''.
\textit{Proceedings of the IEEE/CVF International Conference on Computer Vision}. 
pp. 6869-6879.

\bibitem{yun2019cutmix}
Yun, S., Han, D., Oh, S. J., Chun, S., Choe, J., \& Yoo, Y. (2019).
``Cutmix: Regularization strategy to train strong classifiers with localizable features''.
\textit{Proceedings of the IEEE/CVF International Conference on Computer Vision}
pp. 6023-6032.

\bibitem{barth2018}
Barth R, IJsselmuiden J, Hemming J \& van Henten EJ (2018). 
``Data synthesis methods for semantic segmentation in agriculture: A Capsicum annuum dataset''. 
\textit{Computers and Electronics in Agriculture}. 
144, pp. 284-296.

\bibitem{halstead2018}
Halstead, M., McCool, C., Denman, S., Perez, T., \& Fookes, C. (2018). 
``Fruit quantity and ripeness estimation using a robotic vision system''. 
\textit{IEEE robotics and automation letters}. 
3(4), 2995-3002.

\bibitem{cao2022observation}
Cao, J., Pang, J., Weng, X., Khirodkar, R., \& Kitani, K. (2023). 
``Observation-centric sort: Rethinking sort for robust multi-object tracking''. 
\textit{In Proceedings of the IEEE/CVF conference on computer vision and pattern recognition}. 
pp. 9686-9696.

\bibitem{Bytetrack2022}
Zhang, Y., Sun, P., Jiang, Y., Yu, D., Weng, F., Yuan, Z., ... \& Wang, X. (2022). 
``Bytetrack: Multi-object tracking by associating every detection box''. 
\textit{In European conference on computer vision}. 
pp. 1-21.




\end{thebibliography}
\end{document}